\documentclass[10pt,twocolumn,letterpaper]{article}

\usepackage{cvpr}
\usepackage{times}
\usepackage{epsfig}
\usepackage{graphicx}
\usepackage{amsmath}
\usepackage{amssymb}
\usepackage{amsthm}
\usepackage{bbm}

\usepackage{multirow}

\usepackage[pagebackref=true,breaklinks=true,letterpaper=true,colorlinks,bookmarks=false]{hyperref}

\cvprfinalcopy 

\def\cvprPaperID{****} 

\begin{document}

\title{Light-weight calibrator: a separable component for unsupervised domain adaptation}
\author{Shaokai Ye$^{1}$~~~Kailu Wu$^{2}$~~~Mu Zhou$^{3}$~~~Yunfei Yang$^{2}$~~~Sia huat Tan$^{2}$~~~Kaidi Xu$^{4}$~~~Jiebo Song$^{2}$\\~~~Chenglong Bao$^{2}$\thanks{Corresponding  Authors}\quad~~~ Kaisheng Ma$^{2*}$
\\
    $^1$Institute for interdisciplinary information core technology(IIISCT), China;\\
    $^2$Tsinghua University, China;
    $^3$University of Tsukuba, Japan;
    $^4$Northeastern University, USA\\
}


\maketitle

\begin{abstract}
   Existing domain adaptation methods aim at learning features that can be generalized
   among domains. These methods commonly require to update source classifier to adapt to the target domain and do not properly handle the trade off between the source domain and the target domain.
   In this work, instead of training a classifier to adapt to the target domain, we use a separable component called data calibrator to help the fixed source classifier recover discrimination power in the target domain, while preserving the source domain's performance. When the difference between two domains is small, the source classifier's representation is sufficient to perform well in the target domain and outperforms GAN-based methods in digits. Otherwise, the proposed method can leverage synthetic images generated by GANs to boost performance and achieve state-of-the-art performance in digits datasets and driving scene semantic segmentation. Our method empirically reveals that certain intriguing hints, which can be mitigated by benigh noise similar to adversarial attack to domain discriminators, are one of the sources for performance degradation under the domain shift. 
\end{abstract}

\section{Introduction}
\label{Introduction}
\begin{figure}[htb]

\centering
\includegraphics[scale=0.23]{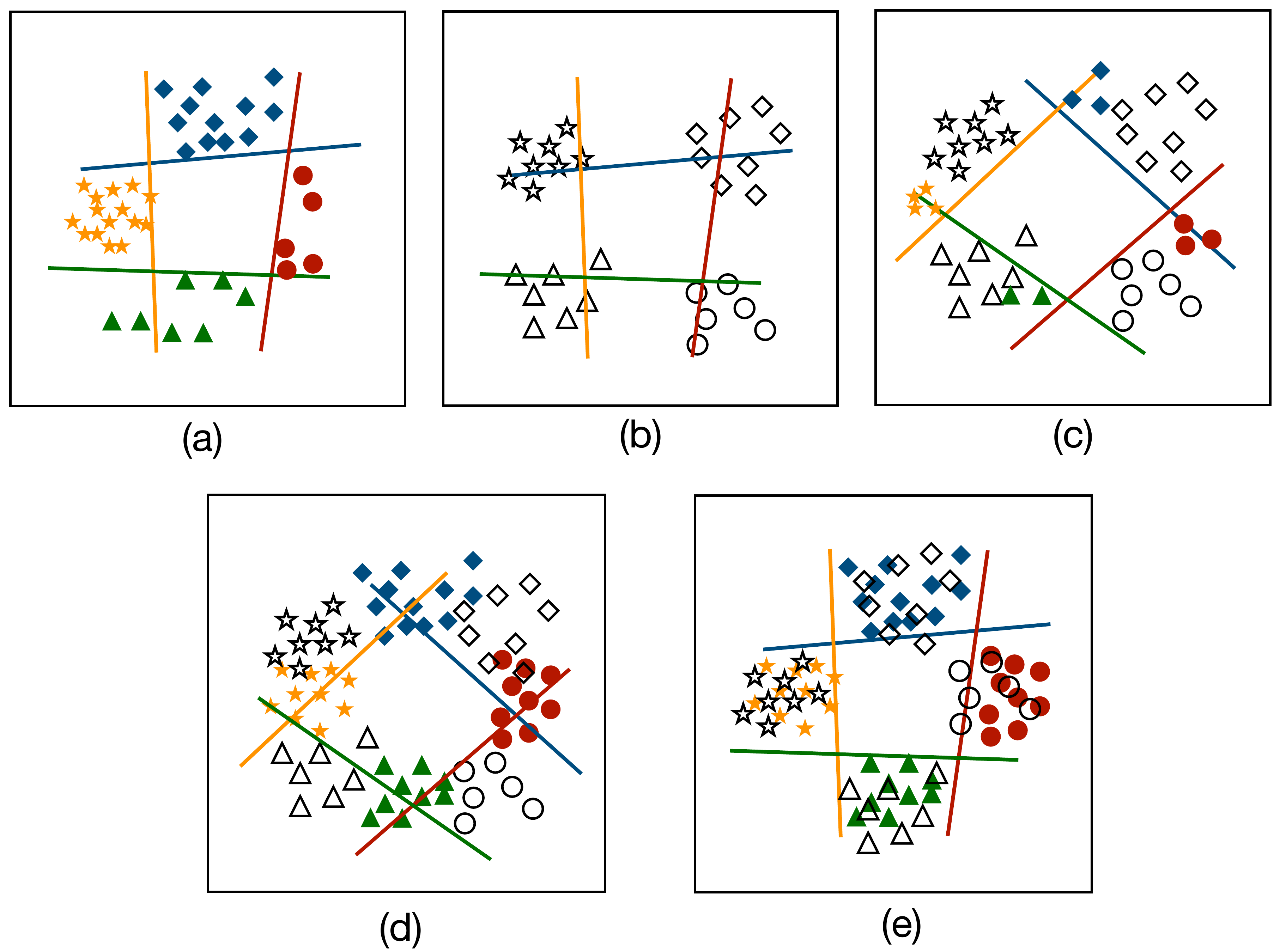}
\caption{\textbf{Concept Illustration}. (a) The source classifier in labeled source domain. (b) The source classifier in unlabeled target domain. (c) Existing methods that are developed to learn domain-invariant features. (d) In real world, the testing set consists of both source domain images and target domain images. (e) The proposed method keeps the representation of source classifier and calibrates target images to fit the source classifier's representation. }
\label{fig:conceptual}
\end{figure}

\begin{figure*}[htb]

\centering
\includegraphics[scale=0.22]{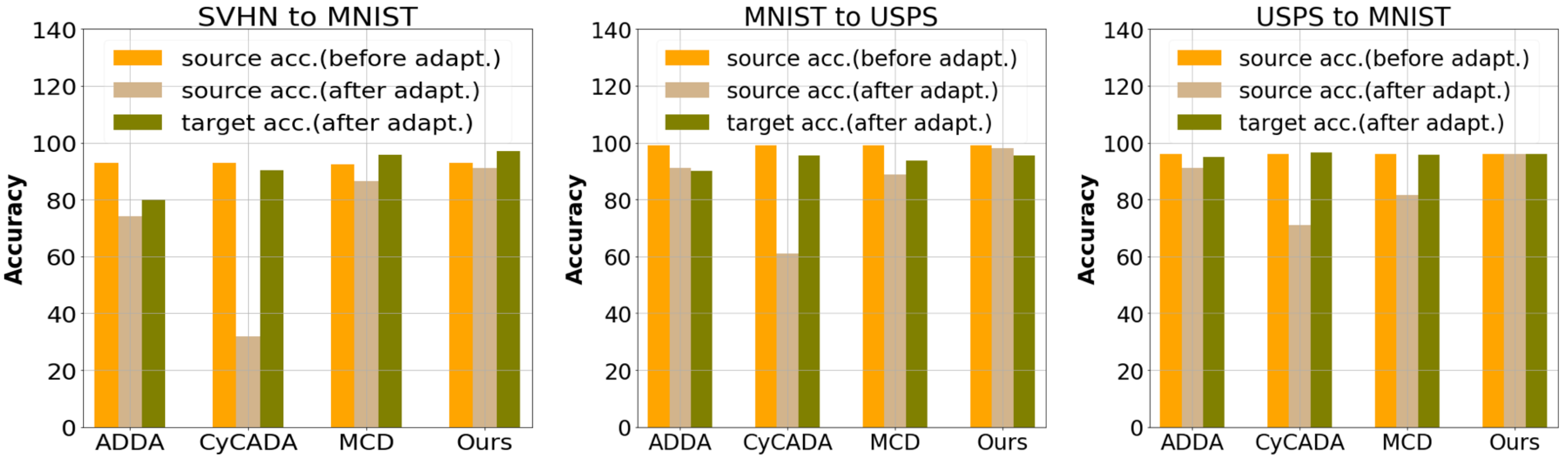}
\caption{\textbf{Performance trade off between source and target domain.}~Some existing methods improve target performance at the expense of source domain performance. On contrast, the proposed method keeps good source domain performance and outperforms these methods in target domain performance.}
\label{fig: weakness}
\end{figure*}

Deep neural networks have achieved great performance in solving diverse machine learning problems. However, solving the so called domain shift problem is challenging when neural networks are trying to generalize across domains~\cite{saenko2010adapting,torralba2011unbiased,recht2019imagenet}. 
Extensive efforts have been made on unsupervised domain adaptation~\cite{saenko2010adapting, ganin2014unsupervised,tzeng2014deep,sun2016deep,hoffman2016fcns,tzeng2017adversarial,hoffman2017cycada,long2018conditional,saito2018maximum}. Early domain adaptation methods use different distance metrics or statistics data to align
neural networks' feature distribution of source domain with their feature distribution of target domain.  Adversarial domain adaptation methods~\cite{ganin2014unsupervised, tzeng2017adversarial} leverage a two players adversarial game to achieve domain adaptation: A domain classifier is encouraged to learn the difference between the feature distribution of two domains while the classification model is encouraged to maximize the classification loss of the domain classifier by learning domain invariant representation that is indistinguishable to the domain classifier. In addition to feature-level adversarial game, there is another line of works that use Generative Adversarial Networks(GANs)~\cite{goodfellow2014generative} to generate source domain images with target domain styles, playing a pixel level adversarial game.

However, there are issues that have been rarely discussed. Consider a neural network that is deployed in a device and the device needs to move between different domains. It moves from a domain that is close to its trained source domain to another domain that has no labeled data. Traditional unsupervised domain adaptation suffices to handle this simple case. However, the devices can freely move to other domains, which include the source domain. This sample but more realistic scenario brings issues to existing methods.  The issues are two folds: (1) Existing methods commonly require to finetune or train a new classifier during domain adaptation. It is not flexible if models are compressed and deployed\cite{han2015learning,zhou2017incremental}. (2) Previous methods omit to show the trade off between source domain performance and target domain performance. Some of them have poor performance trade off as indicated in Figure~\ref{fig: weakness}. Therefore, when the environments are constantly changing, existing methods are likely to have performance degradation and are not able to adapt to new environments in a flexible way.

Some prior works try to work on changing domains~\cite{wulfmeier2018incremental,bobu2018adapting}.   Bobu~\etal~\cite{bobu2018adapting} proposes to adapt to continuously changing target domains and Wulfmeier~\etal~\cite{wulfmeier2018incremental} proposes to incrementally adapt to changing domains. However their methods require to finetune the model, and after the model is deployed, the method cannot work properly for unanticipated new domains. We thereby propose two properties a domain adaptation method should have for changing target domains with deployed models.

\textit{(1)~Good trade-off between source and target domain.} Given the complexity of the real world, it is unrealistic to assume that the one chosen target domain is the ultimate application domain. Existing methods assume that the source domain only consists of synthetic images and omit to show the source domain performance after domain adaptation, mostly because that it is assumed the source domain will not be encountered again. A counter example is that both source domain and target domain consist of real world images and source domain will also be encountered. In this case, sacrificing source domain performance is not acceptable.

\begin{figure}[htb]
\centering
\includegraphics[scale=0.15]{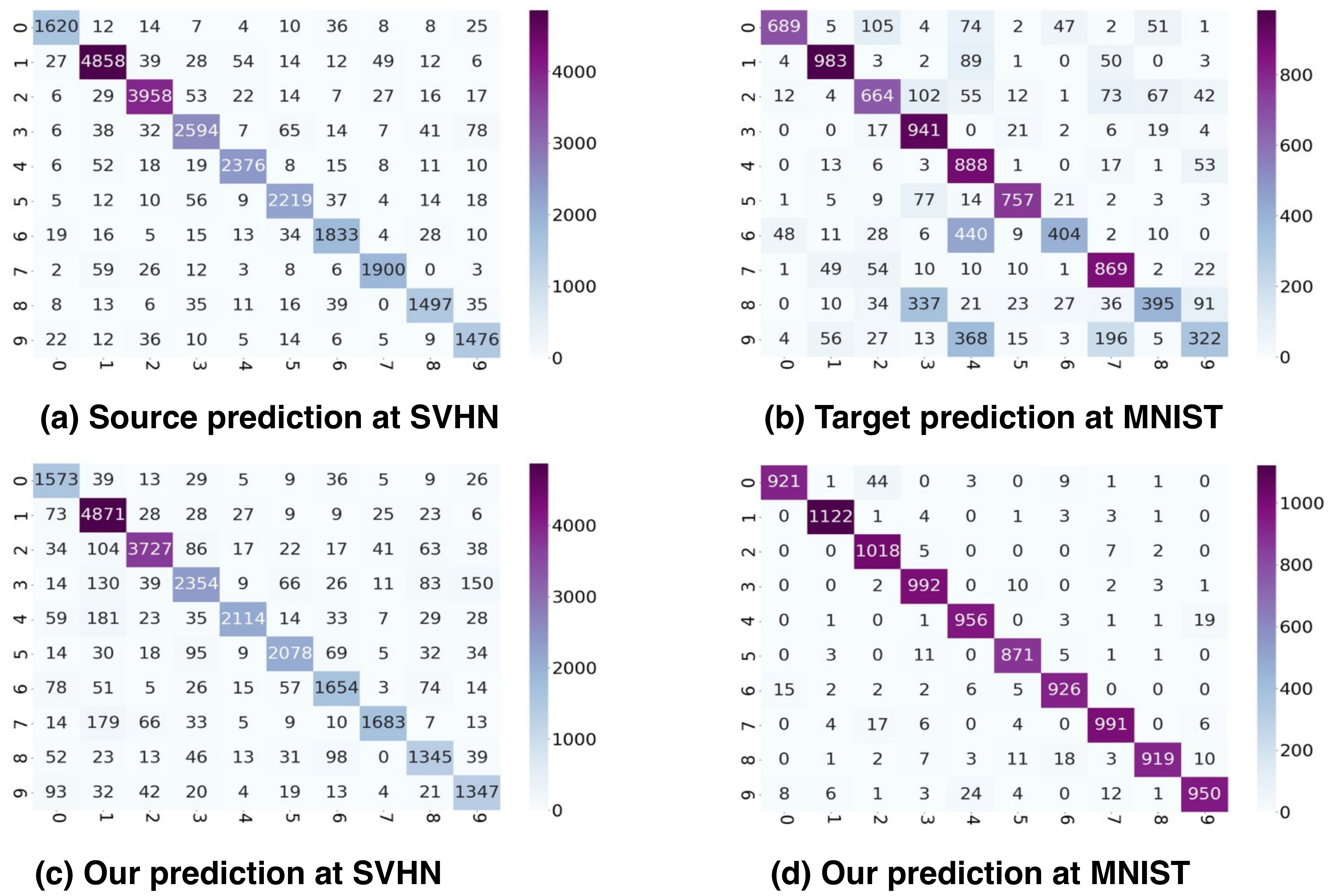}

\caption{SVHN to MNIST task. Source classifier LeNet is trained in SVHN. (a) The source classifier's prediction on SVHN. (b) The source classifier's prediction on MNIST. (c) The source classifier's prediction on SVHN, with data calibrator. (d) The source classifier's prediction on MNIST, with data calibrator.
}
\label{fig:confusion_matrix}
\end{figure}

\textit{(2)~Flexibility to adapt to arbitrary new domains after being deployed.}
Deep neural networks are widely deployed in specialized devices~\cite{han2016eie}. Usually, they are compressed via model compression methods~\cite{han2015learning,zhou2017incremental,zhang2018systematic,Ye_2019_ICCV,ren2019admm} before being deployed and they are not expected to be updated after being deployed. As far as we know, all existing domain adaptation methods will require finetune the models, which contradicts with model compression methods.


It is natural to expect that collecting more data will make a neural network learn universal representation and tremendous investment is made for collecting bigger datasets~\cite{deng2009large,kuznetsova2018open}. However, datasets are found to contain database bias~\cite{torralba2011unbiased,recht2019imagenet}. Training against large datasets does not guarantee the performance of models under changing environments. Therefore, adapting to unanticipated new environments will be necessary and lacking of the flexibility will be an issue.

In this work, we take the first step
to mitigate both limitations and formulate unsupervised domain adaptation in a novel way. Figure~\ref{fig:conceptual} illustrates the difference between previous methods and our method in the conceptual level. Previous methods commonly update the source classifier's weights when domain adaptation is needed while ours modifies inputs to achieve domain adaptation. 

We refer existing methods that attempt to learn cross-domain models as monolithic domain adaptation approach. On contrast, we propose a separable component called data calibrator to achieve domain adaptation, which can be seen as a distributed domain adaptation approach. In our framework, the source classifier is responsible for learning representation under supervised training and the data calibrator is responsible for achieving domain adaptation via unsupervised training. 

Our core observation is that the learnt representation from the source domain is not as bad as we thought as shown in Figure~\ref{fig:confusion_matrix}. The performance degradation brought by domain shift can be mitigated by slightly modifying the target domain images by adding perturbation , which we refer as calibration, to the images. By applying calibration to target domain images, these images fit the source classifier's learnt representation significantly better. We show that we can train a light-weight data calibrator whose number of parameters is only 0.25\% to 5.8\% of the deployed model and we can use it to adapt the deployed model to arbitrary target domains.

We also want to emphasize that our study focus on the setting that the source domain and the target domain share the common label space otherwise the source classifier will not work properly in the target domain.

To summarize our contributions:
\begin{itemize}
    \item We propose a data calibrator to calibrate target domain images to better fit source classifier's representation while maintaining the source domain performance. We improve previous state-of-the-art average accuracy from 95.1\% to 97.6\% in digits experiments and frequency weighted IoU from 72.4\% to 75.1\% in GTA5 to CityScapes adaptation. 
    \item The proposed data calibrator is light weight and can be less than 1\% in terms of number of parameters compared to the deployed model in GTA5 to CityScapes adaptation and it is a separable domain adaptation approach for it does not need to update the source classifier's weights, thus very convenient for deployment.
    \item We give new insights on what causes the performance degradation under domain shift and show how to counter it correspondingly.
\end{itemize}

\section{Related Work}
\label{Related Work}

\textbf{Unsupervised Domain Adaptation}
Visual domain adaptation can trace back to~\cite{saenko2010adapting}. Early domain adaptation methods focus on aligning deep representation between two domains by using Maximum Mean Discrepancy(MMD)~\cite{quinonero2008covariate,tzeng2014deep,long2015learning} whereas deep Correlation Alignment (CORAL)~\cite{sun2016deep} used statistics such as mean and covariance to achieve feature alignment. 

Another line of works leverage the idea of domain classifiers. Torralba~\etal~\cite{torralba2011unbiased} used "name the database" to demonstrate that databases are commonly biased and it is even possible to train a domain classifier to correctly classify images to databases they come from. Intuitively, if a domain classifier can learn the difference between source domain and target domain from pixels, then it is also possible for a domain classifier to learn the difference between deep representation of source domain images and target domain images. A line of works explore the idea of training a classifier that confuses the domain classifier by maximizing the domain confusion loss~\cite{tzeng2015simultaneous,ganin2014unsupervised, tzeng2017adversarial,ganin2016domain,sun2016return,tzeng2014deep}.
In addition to the attempt of confusing a domain classifier in the feature level, pixel level adaptation is also explored. Hoffman~\etal~\cite{hoffman2016fcns} achieves pixel level adaptation for segmentation task, but it uses neural networks' hidden layer output for pixel level adaptation. Our method incorporates both pixel level domain classifier and feature level domain classifier. The pixel level classifier we use directly takes the pixels as inputs, closer to the spirit of "name the dataset"~\cite{torralba2011unbiased}.

\textbf{Generative Adversarial Networks}
Another line of works leverages the power of Generative Adversarial Networks (GANs)~\cite{goodfellow2014generative} to generate source images with target images' style.
The first of this kind is CoGANs~\cite{liu2016coupled} that jointly learns the source domain representation and the target domain representation by forcing the weight sharing between two GANs. Bousmalis~\etal~\cite{bousmalis2017unsupervised} used GANs to produce images that have similar styles to target domain and makes the target task classifier to train images of both. Hoffman~\etal~\cite{hoffman2017cycada} proposes to use semantic consistency loss and cycle consistency loss and achieve significantly better domain adaptation performance. As a comparison, our method can outperform those methods without requiring high-resources to train GANs.

\textbf{Adversarial Attack}
Neural networks are known for suffering from adversarial attacks~\cite{szegedy2013intriguing,xu2018structured,xu2019topology,zhao2019design}. The simplest form of adversarial attack is FGSM~\cite{goodfellow2014explaining}, which adds a calculated perturbation on the original image, making neural networks misclassify with high confidence. Interestingly, the proposed data calibrator also uses an additive perturbation on images to achieve domain adaptation. The connection between adversarial attack and domain adaptation will be revealed at the objective function in our framework. Essentially, our data calibrator learns to generate adversarial examples that maximize classification loss of domain classifiers.Recently, Ilyas~\etal~\cite{ilyas2019adversarial} demonstrates that adversarial attack might leverage "non-robust features" to control classifiers' prediction. We believe that "non-robust features" play an important role in performance degradation brought by domain shift. We will provide more analysis about the connection between our method and adversarial attack in Section \ref{discussion}.

\section{A Separable Calibrator For Unsupervised Domain Adaptation}
\label{Methodology}
\begin{figure*}[htb]

\centering
\includegraphics[scale=0.37]{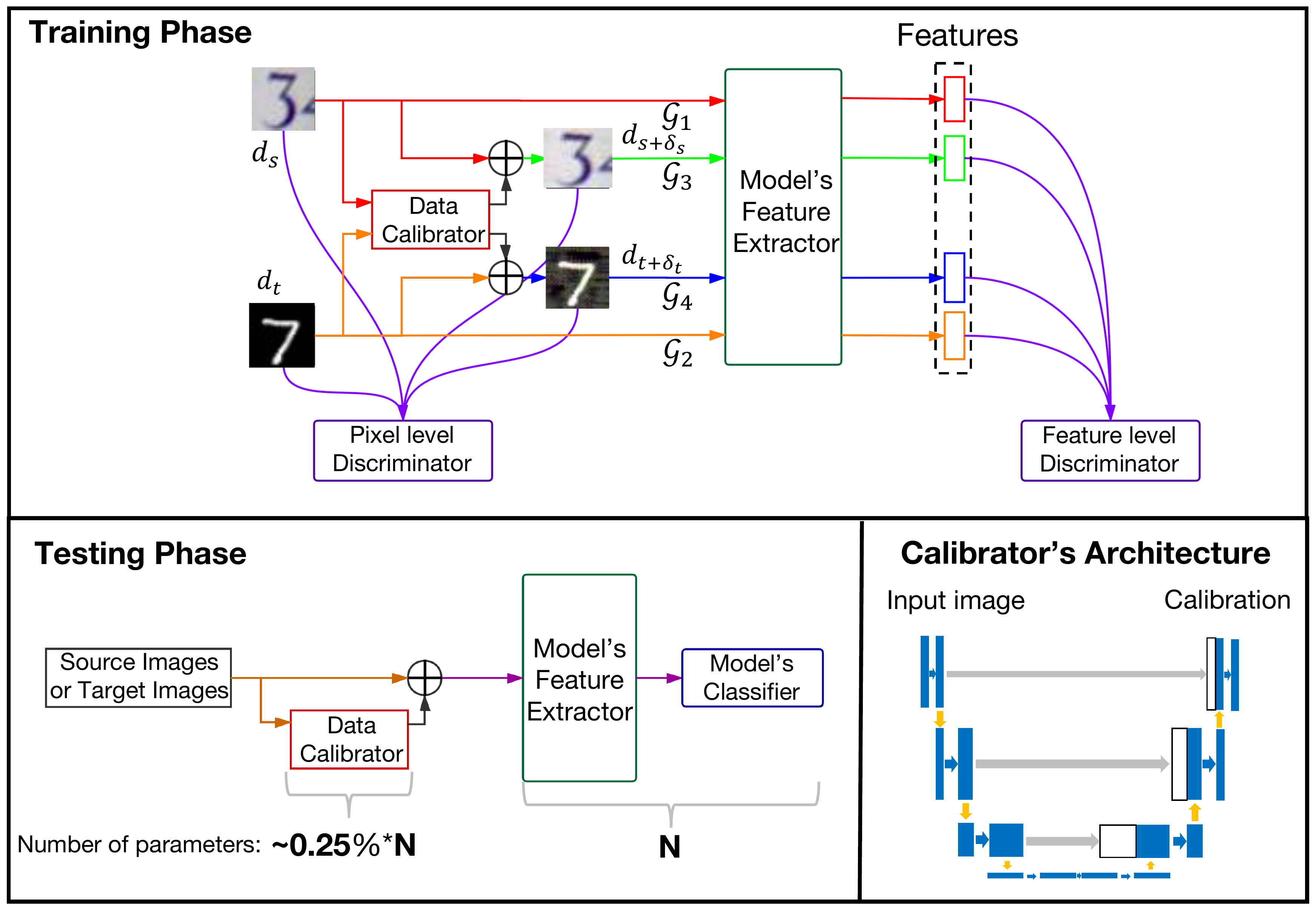}

\caption{\textbf{Training, testing phase and data calibrator architecture.} In the training phase, the pixel level discriminators and the feature space discriminator try to discriminate images to 4 groups while the data calibrator tries to fool both discriminators to treat calibrated images as the source images. In the testing phase, the deployed model takes calibrated images as inputs. The architecture for the data calibrator consists of down sampling layers, up sampling layers and skip connections.\label{fig:framework}}
\end{figure*}

\subsection{The overview of the method}
In unsupervised domain adaptation, we have access to source domain images $X_s$ and labels $Y_s$ drawn from the source domain distribution $p_s(x,y)$, and target domain images $X_t$ drawn from a target domain distribution $p_t(x,y)$, where there are no labels. Let $F_s$ be the learned classifier for source domain images. The goal of our work is to design a data calibrator $G_c$ such that $F_s\circ G_c$ achieves high accuracy on both source and target domain data. As the classifier $F_s$ is only trained on source domain and there is no information related to the target, the data calibrator $G_c$ has to satisfy:
\begin{equation}\label{req:calibrator}
\begin{aligned}
    F_s(G_c(X_t))\sim F_s(X_s),&\quad F_s(G_c(X_s))\sim F_s(X_s)
\end{aligned}
\end{equation}
where $X_t$ and $X_s$ are from target and source domain respectively.

Let $F_s=C_s\circ M_s$ where $M_s$ the feature extractor and $C_s$ is the final classifier. 
A relaxed condition for achieving \eqref{req:calibrator} is to impose the Lipschitz condition on $F_s\circ G_c$, i.e.
\begin{equation*}
    \|F_s\circ G_c (x)-F_s\circ G_c(y)\|\leq L\|x-y\|,
\end{equation*}
for some constant $L>0$ which is a stability condition. Therefore,
the following two constraints are imposed on the data calibrator: 
\begin{equation}\label{req:calibrator_2}
\begin{aligned}
G_c(X_t)\sim X_s,&\quad G_c(X_s)\sim X_s\\
M_s(G_c(X_t))\sim M_s(X_s),&\quad M_s(G_c(X_s))\sim M_s(X_s) 
\end{aligned}
\end{equation}
It is noted that $G_c(x)$ denotes the input of $F_s$ and $M_s$ denotes the feature map which implies the alignment on both pixel and feature level for source and target domain data. This motivates the following loss function:
\begin{equation}
  \begin{aligned}
\min _{G_c}~ & H(X_s || G_c (X_t)) + H(M_s(X_s)||M_s(G_c(X_t)))\\
& H(X_s||G_c(X_s)) + H(M_s(X_s)||M_s(G_c(X_s))),
\label{OPT}
\end{aligned}  
\end{equation}
where $H$ denotes the Cross entropy. The loss function in \eqref{OPT} encourages the data calibrator for domain adaption while keeping the performance in source domain. In this work, the data calibrator is set as $G_c=I+G_c^{'}$, i.e. only the perturbation is learned by the calibrator. However, as the target information is blind, minimizing \eqref{OPT} is difficult and another method is needed for training the calibrator $G_c$.

\subsection{Adversarial Domain Adaptation with Proposed Calibrator}
In this work, we extend the traditional adversarial domain adaption methods~\cite{ganin2014unsupervised,ganin2016domain,tzeng2017adversarial} and train the proposed calibrator via adversarial learning instead of minimizing~\eqref{OPT}. 

Traditional adversarial domain adaptation methods play a adversarial game between the target classifier $F_t$ and feature discriminator $D_{feat}$. Because they update weight parameters of $F_t$ to maximize the confusion loss of domain discriminators, the resulted adapted models lack the flexibility of adjusting to new domains after being deployed and are under the risk of sacrificing source domain performance. 

On contrast, the basic idea of our extended adversarial domain adaption method is that let there be pixel level domain discriminator $D_{pixel}$ and feature level domain discriminator $D_{feat}$. And let a data calibrator modify images such that domain discriminators $D_{pixel}$ can no longer distinguish between $G_c(X_t)$ and $X_s$ nor between $G_c(X_s)$ and $X_s$. Meanwhile, the corresponding features of calibrated images are also confusing $D_{feat}$ such that the feature level discriminator can no longer distinguish between $M_s(G_c(X_s))$ and $M_s(X_s)$ nor between $M_s(G_c(X_t))$ and $M_s(X_s)$. After the calibrator is trained, inputs are fed to the calibrator before fed to the model, as shown in the testing phase at Figure~\ref{fig:framework}.


As shown in Figure~\ref{fig:framework}, the training of the proposed method needs a trained source classifier $F_s$. Let the source classifier $F_s$ be trained by the following loss function:


\begin{align}\label{classification}
\mathcal{L}_{\mathrm{source}}\left(f_{S}, X_{S}, Y_{S}\right)=-\mathbb{E}_{\left(x_{s}, y_{s}\right) \sim\left(X_{S}, Y_{S}\right)}   \notag \\
\sum_{k=1}^{K} \mathbbm{1}_{\left[k=y_{s}\right]} \log \left(\sigma\left(f_{S}^{(k)}\left(x_{s}\right)\right)\right).
\end{align}

Based on the learned classifier $F_s$, the pixel level domain discriminator $D_{pixel}$ and feature level domain discriminator $D_{feat}$ are proposed for training the calibrator such that the pixel and feature level alignment conditions \eqref{req:calibrator_2} is satisfied. Furthermore, in order to have a finer discrimination power among images and features from source domain and target domain, we divide the inputs of the domain discriminators into 4 groups inspired by the few shot domain adaptation~\cite{motiian2017few}.

These four groups ($\mathcal{G}_{i}$,i=1,2,3,4) are defined as as follows: $\mathcal{G}_{1}$ represents source domain images $X_s$, $\mathcal{G}_{2}$ represents target domain images $X_t$. Therefore, learning to distinguish images and features from $\mathcal{G}_{1}$ and $\mathcal{G}_{2}$ encourages the domain discriminators to learn the distributions of source domain and target domain. Additionally, calibrated source images $G_c(X_s)$ are defined to belong to $\mathcal{G}_{3}$ and calibrated target images $G_c(X_t)$ are defined to belong to $\mathcal{G}_{4}$ as to provide learning signal for the adversarial game. Let $y_{\mathcal{G}_{i}},i=1,2,3,4$ be the group labels for each group.

\textbf{Feature Level Discriminator.}~The feature level discriminator aims to discriminate feature level distribution $M_s(\mathcal{G}_{i})$. Its objective is to minimize categorical cross entropy loss as following:

\begin{align}
\mathcal{L}_{feat-D}=-E\left[\sum_{i=1}^{4} y_{\mathcal{G}_{i}} \log (D_{feat}(M(\mathcal{G}_{i})))\right],
\end{align}
In our work, the feature level discriminator $D_{feat}$ is a simple neural network with only two fully connected layers. During training, the feature level discriminator learns to discriminate features distribution of $M_s(\mathcal{G}_{i})$. 

\textbf{Pixel Level Discriminator.}~The limitation of using only feature level discriminator is that feature level discriminator cannot fully capture the information in the pixel level after images are transformed via pooling layers and strided convolutional layers of the model. Thus, following the original idea of~\cite{torralba2011unbiased}, a pixel level discriminator $D_{pixel}$ is added to learn pixel level distribution of $\mathcal{G}_{i}$ by following objective function:
\begin{align}
\mathcal{L}_{pixel-D}=-E\left[\sum_{i=1}^{4} y_{\mathcal{G}_{i}} \log (D_{pixel}(\mathcal{G}_{i}))\right].
\end{align}
The pixel level discriminator $D_{pixel}$ shares the same architecture as the feature level discriminator $D_{feat}$, i.e.\ a two layer fully connected network. The biggest challenge for the pixel level discriminator $D_{pixel}$ is 
its tendency of over-fitting to the training set. From our observations in experiments, the validation accuracy starts going down when the training loss for the pixel discriminator gets very low,. Indeed, if the calibrator is optimized towards to a pixel level discriminator that overfits, it looses the generalization power. Therefore, we apply following tricks to the inputs of pixel level discriminator to prevent it from overfitting: (1) A image patch is randomly taken from the image. (2) The pixels of the patch is randomly shuffled in the spatial axis. By applying the above two tricks, the overfitting is mitigated.

\textbf{Data Calibrator.} The data calibrator's goal is to fool both the pixel level discriminator $D_{pixel}$ and feature level discriminator $D_{pixel}$ by the following loss function:
\begin{align}
\mathcal{L}_{Calibrator}=-E[y_{\mathcal{G}_{1}} \log (D_{feat}(M_s(\mathcal{G}_{3})))\notag\\+y_{\mathcal{G}_{1}} \log (D_{feat}(M_s(\mathcal{G}_{4})))\notag\\+y_{\mathcal{G}_{1}} \log (D_{pixel}(\mathcal{G}_{3}))\notag\\+y_{\mathcal{G}_{1}} \log (D_{pixel}(\mathcal{G}_{4}))],
\end{align}
from which the learned calibrator is expected to learn knowledge in source and target domain and satisfies \eqref{req:calibrator_2}. 
\begin{table*}[h!]
\begin{center}
\begin{tabular}{lcccc}
\hline
\textbf{Method} & \textbf{MNIST to USPS} & \textbf{USPS to MNIST} & \textbf{SVHN to MNIST} & \textbf{Average Acc.}\\  \hline
ADDN~\cite{tzeng2017adversarial}        &  90.1                  &  95.2                  &  80.1       &88.5           \\
CoGAN~\cite{liu2016coupled}         &  91.2               &  89.1               & -           & -           \\
SBADA~\cite{russo2018source}  & \textbf{97.6} & 95.0 & 76.1 & 89.6
\\
CYCADA~\cite{hoffman2017cycada}        &  95.6               &  96.5               &  90.4      & 94.2         \\
CDAN~\cite{long2018conditional}          &  95.6                &  98.0     &89.2   &94.3
\\
PFA~\cite{chen2019progressive}     & 95.0                & -                  &93.9    & -
\\

MSTN~\cite{xie2018learning}   & 92.9 & 97.6 & 93.3 & 94.6
\\
MCD~\cite{saito2018maximum}   & 93.8 & 95.7 & 95.8 & 95.1
\\

\hline
Ours            &  95.6               &  97.1                  &  97.1           & 96.6       \\ 
CyCleGAN+Ours            &  97.1               &  \textbf{98.3}                  &  \textbf{97.5}   & \textbf{97.6}               \\ \hline

\end{tabular}
\caption{\textbf{Results on digits datasets for unsupervised domain adaptation}. Our method achieves state-of-the-art performance without using stylized source images. Our method can be further improved by using stylized source images.\label{exp:digits}}
\end{center}
\vspace{-5mm}

\end{table*}
The total training loss of our data calibrator can be divided into two parts.
When the calibrator tries to fool domain discriminators to treat $\mathcal{G}_{3}$ as $\mathcal{G}_{1}$, the calibrator tends to approximate the identity mapping. On contrast, when the calibrator tries to fool domain discriminators to treat $\mathcal{G}_{4}$ as $\mathcal{G}_{1}$, the calibrator is to calibrate target domain images to mitigate the domain shift.

The ResNet generator~\cite{johnson2016perceptual} is used as the architecture of the calibrator for digits and GTA5 to CityScapes experiments. It consists of downsampling layers, upsampling layers and skip connections, as shown in Figure~\ref{fig:framework}. It is noted that the performance does not simply get better when the calibrator network is getting larger. However, reducing the width can improve training as it is believed that it prevents the data calibrator from overfitting when the training data is not sufficient.
Additionally, applying $L_{\infty}$ norm constrain to the output of the data calibrator plays an important role in GTA5 to CityScapes adaptation. We will give a more detailed discussion on this constrain in Section~\ref{discussion}.

\section{Evaluation and Results}
\label{evaluation}

\begin{figure*}[htb]

\centering
\includegraphics[scale=0.24]{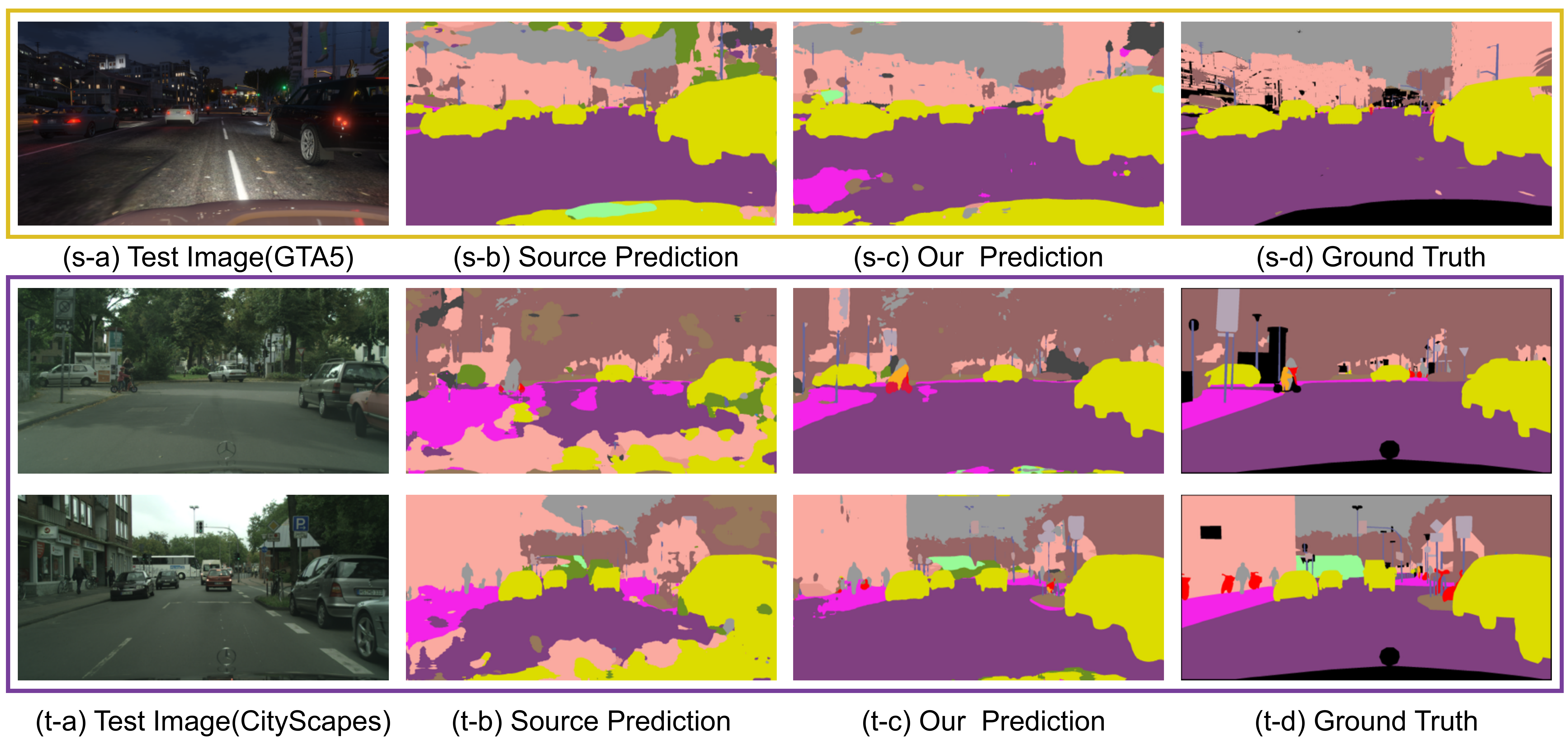}

\caption{Semantic Segmentation results for GTA5 to CityScapes. (s-a) Test images from GTA5. (s-b) Predictions from the model trained in GTA5. (s-c) Our prediction. (s-d) Ground truth annotations for test images.
(t-a) Test images from CityScapes. (t-b) Predictions from the model trained in GTA5. (t-c) Predictions from our method. (t-d) Ground truth annotations for test images.}
\label{fig: segmentation}
\end{figure*}

\begin{table*}[htb]

\centering
\setlength\tabcolsep{1.3pt}
\begin{tabular}{ccccccccccccccccccccccc}
\hline
 & {\rotatebox{90}{road}} & {\rotatebox{90}{sidewalk}} & {\rotatebox{90}{building}} & {\rotatebox{90}{wall}} & {\rotatebox{90}{fence}} & {\rotatebox{90}{pole}} & {\rotatebox{90}{traffic light}} & {\rotatebox{90}{traffic sign}} & {\rotatebox{90}{vegetation}} & {\rotatebox{90}{terrain}} & {\rotatebox{90}{sky}} & {\rotatebox{90}{person}} & {\rotatebox{90}{rider}} & {\rotatebox{90}{car}} & {\rotatebox{90}{truck}} & {\rotatebox{90}{bus}} & {\rotatebox{90}{train}} & {\rotatebox{90}{motorbike}} & {\rotatebox{90}{bicycle}} & {\rotatebox{90}{\textbf{mIoU}}} & {\rotatebox{90}{\textbf{fwIoU}}} & {\rotatebox{90}{\textbf{Pixel acc.}}} \\ \hline
\multicolumn{1}{c|}{Source only}   & 42.7 & 26.3 & 51.7 & 5.5 & 6.8 & 13.8 & 23.6 & 6.9 & 75.5 & 11.5 & 36.8 & 49.3 & 0.9 & 46.7 & 3.4 & 5.0 & 0.0 & 5.0 & 1.4 & 21.7 & 47.4 & 62.5 \\
\multicolumn{1}{c|}{CyCADA}   & 79.1 & 33.1 & 77.9 & 23.4 & 17.3 & 32.1 & 33.3 & 31.8 & 81.5 & 26.7 & 69.0 & 62.8 & 14.7 & 74.5 & 20.9 & 25.6 & 6.9 & 18.8 & 20.4 & 39.5 & 72.4 & 82.3 \\
\multicolumn{1}{c|}{Ours}   & \textbf{83.5} & \textbf{35.2} & \textbf{79.9} & \textbf{24.6} & 16.2 & \textbf{32.8} & 33.1 & 31.8 & \textbf{81.7} & \textbf{29.2} & 66.3 & \textbf{63.0} & 14.3 & \textbf{81.8} & \textbf{21.0} & \textbf{26.5} & \textbf{8.5} & 16.7 & \textbf{24.0} & \textbf{40.5} & \textbf{75.1} & \textbf{84.0} \\ \hline
\multicolumn{1}{c|}{Target} & 97.3 & 79.8 & 88.6 & 32.5 & 48.2 & 56.3 & 63.6 & 73.3 & 89.0 & 58.9 & 93.0 & 78.2 & 55.2 & 92.2 & 45.0 & 67.3 & 39.6 & 49.9 & 73.6 & 67.4 & 89.6 & 94.3 \\ \hline
\end{tabular}
\caption{\textbf{Adaptation between GTA5 and CityScapes}. Source only shows results of DRN-26~\cite{yu2017dilated} trained in GTA5 and tested in CityScapes. Target only shows results of DRN-26 trained in CityScapes and tested in CityScapes. Our method outperforms CyCADA in mean IoU, freqency weighted IoU and pixel accuracy. In particular, our frequency weighted IoU is 2.7\% better than CyCADA.}
\label{exp:gta5}
\end{table*}

In this section, we evaluate our method under unsupervised domain adaptation setting on digits and driving scene semantic segmentation tasks.

\textbf{Digits}  We evaluate our method on three commonly used digits datasets: MNIST~\cite{lecun1998gradient}, USPS, and SVHN~\cite{netzer2011reading}. We use the same data processing and LeNet architecture as Hoffman~\etal~\cite{hoffman2017cycada} and perform three unsupervised domain adaptation tasks: USPS to MNIST, MNIST to USPS and SVHN to MNIST. We report our results of using unstylized source images and stylized source images produced by CycleGAN~\cite{zhu2017unpaired} respectively.

\textbf{GTA5 to CityScapes} GTA5~\cite{richter2016playing} is a synthetic driving scene dataset and CityScapes~\cite{cordts2016cityscapes} is a real world driving scene dataset. The GTA5 dataset has 24966 densely labeled RGB images of size $1914\times1052 $, which contains 19 common classes with CityScapes, as we included in Table~\ref{exp:gta5}. The CityScapes dataset contains 5000 densely labeled RGB images of size $2040\times1016$ from 27 cities. In this work, we use DRN-26~\cite{yu2017dilated} as the source classifier. We use the released DRN-26 model from CyCADA~\cite{hoffman2017cycada} as our source classifier, which is trained in stylized GTA5 images.

All components are implemented using Pytorch. For digits experiments, source classifiers and other components are trained with the Adam optimizer with learning rate 1e-4. We use batches of 128 samples from each domain and the images are zero-centered and rescaled to $[-1,1]$. For GTA5 to CityScapes experiments, we use Adam optimizer with learning rate 1e-4 with batch size 6. We use same LeNet architecture as CyCADA for all digits experiments and DRN-26~\cite{yu2017dilated} for GTA5 to CityScapes task. Our best results are obtained within 50 epochs for digits and within 10 epochs for GTA5 to CityScapes. 

Details about other components such as architecture of the data calibrator and domain discriminators can be found at Appendix.

\subsection{Digits Experiments}
As we show in Figure~\ref{fig:confusion_matrix}, the learnt representation of source classifier is not as bad as we thought. To prove that, we show that without training a new classifier or using stylized source images produced by GANs, we can just use the source classifier trained in the source domain and train a data calibrator to modify the images to fit the source classifier's representation. As we show in Table \ref{exp:digits}, using data calibrator alone can outperform previous methods in average accuracy. For difficult task such as SVHN to MNIST, we can further boost our performance by using stylized source images~\cite{zhu2017unpaired} as source domain, resulting in 7\% performance improvement compared to CyCADA, another method that leverages stylized source images for unsupervised domain adaptation.

\subsection{Performance Trade off Among Domains}
As we discuss in Section~\ref{Introduction}, existing methods omit to show the trade off between source domain performance and target domain performance. In this subsection, we show that many existing methods have poor source and target domain performance trade off. We use the released code from CyCADA~\cite{hoffman2017cycada},ADDA~\cite{tzeng2017adversarial} and MCD~\cite{saito2018maximum}, follow their setting and train their adapted models to get similar reported target domain performance. We then test their adapted model on the source domain and target domain, report the performance before domain adaptation, after domain adaptation. We observe from Figure~\ref{fig: weakness} that, while ADDA has close performance at USPS to MNIST as ours in the target domain, but its source domain performance is 5\% lower than ours. CyCADA has a lot higher target domain performance compared to ADDA, however, it sacrifices source domain performance significantly. MCD is better than the other two in performance trade off, but it uses a baseline that has over-parameterized fully connected layers and does not converge well when we replace their backbone with the same LeNet architecture other approaches and ours use. While our method can be further improved by using GAN generated images as source domain, using the data calibrator alone without stylized images can already surpass these methods in both source domain performance and target domain performance as indicated by Figure~\ref{fig: weakness}. 

\subsection{GTA5 to Cityscapes}
GTA5 to Cityscapes is a unsupervised domain adaptation task that is closer to real world setting. Compared to classification task, segmentation task is more challenging because that finer domain adaptation methods are required to mitigate domain shift in pixel levels.

As shown in Table~\ref{exp:digits}, our method has better results in all three commonly used metrics such as mIoU, fwIoU, and pixel accuracy. In particular, our fwIoU is 2.7\% better than CyCADA. In Figure \ref{fig: segmentation}, we visualize our semantic segmentation results. From (s-b) to two rows at (t-b), we observe the performance degradation brought by the domain shift. (s-c) and (t-c) shows the segmentation results produced by our method. Our method largely mitigates the performance degradation in target domain as well as maintaining source domain performance. Because we improve the accuracy of cars by a large margin, the visualization for cars are quite close to the ground truth annotations.

\begin{figure*}[htb]
\centering
\includegraphics[scale=0.3]{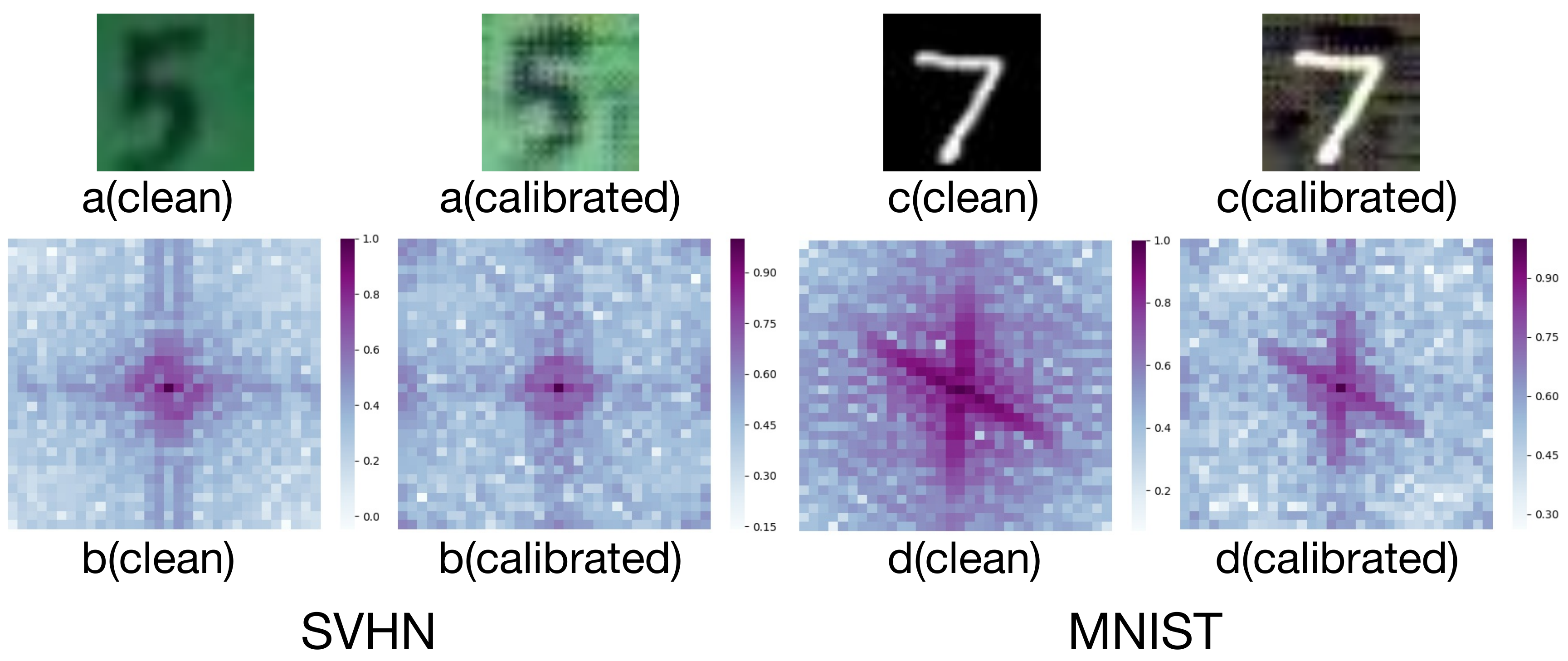}
\caption{\textbf{Images from SVHN to MNIST adaptation} Images before and after being calibrated and their view in the frequency domain. The appearance of images are not changed much unlike what style transfer GANs do. In frequency domain, high frequency information is reduced.
}
\label{fig:frequency}
\end{figure*}

\section{Discussion}\label{discussion}
This section is organized as following: Section 5.1 focuses on the analysis of calibrated images in the frequency domain. In Section 5.2, we discuss the connection between adversarial attack and domain adaptation. In Section 5.3, deployment of the data calibrator will be discussed. 

\subsection{Fourier Perspective}
We use Fast Fourier Transform(FFT) to show images before and after adding calibration. It can be seen in Figure~\ref{fig:frequency} that the high frequency information is decreased after images are added with the output of our data calibrator. High frequency information is often related to textures that varies significantly across domains. Yin~\etal~\cite{yin2019fourier} demonstrates that naturally trained models are biased towards high frequency information, which makes models suffer from high frequency noise. Our method might help remove these high frequency information from images thus mitigating the domain shift problem.

\subsection{Connection to Adversarial Attack}

Compared to other methods that train classifiers to adapt to target domains, in our domain adaptation framework, once trained in the source domain, the source classifier is not updated and we fully rely on the representation learnt in the source domain to perform tasks in the target domain. Thus the additive calibration produced by our data calibrator needs to figure out how to transform target domain images to a form that better fits the source classifier's representation.

But what does it mean by modifying target domain images to better fit the source classifier's representation? We first hypothesize that there are two candidate explanations of what the data calibrator does: (1) the data calibrator acts as a style transfer GAN that converts the style of target domain images to source domain images's thus achieve domain adaptation. (2) the data calibrator learns to manipulate non-robust features that are useful to neural networks but are intriguing to human~\cite{ilyas2019adversarial}. Our data calibrator might learn to suppress these non-robust features thus mitigate the issue brought by the domain shift.

As can be observed from Figure~\ref{fig:frequency}, the images modified by our calibrator do not change their appearance in the way the style transfer GAN usually does. We also follow the convention of adversarial attack~\cite{goodfellow2014explaining} to limit $L_{\infty}$ of the calibration and provide the plot in Appendix. Our best result in Table~\ref{exp:gta5} is obtained by limiting the $L_{\infty}$ of calibration to 0.01, so small that a human might not be able to tell. Essentially, our data calibrator is trained to produce a perturbation that fools the domain discriminators with human imperceivbale perturbation, which is very similar to the behavior of adversarial attacks ~\cite{szegedy2013intriguing,goodfellow2014explaining}.
This suggests that our data calibrator is not performing style transfer but leveraging intriguing hints to mitigate the domain shift problem. Our method suggests that there is a potential connection between adversarial attack and domain adaptation and our results should be interesting to both research community.

\subsection{Calibrator for Deployment}

As we discuss in Section~\ref{Introduction}, one of the limitations of existing domain adaptation methods is the lack of flexibility. As far as we know, most existing domain adaptation methods will require finetune the deployed model when there is a new target domain. However, the deployed model is usually compressed and stored in specialized hardwares thus adapting the deployed models to new domains requires a long, costly process and might not be fast enough for time-sensitive applications. 

On contrast, our method does not require updating the deployed model and has greater flexibility when adapting to a new domain is desired. Additionally, the overhead brought by the calibrator is moderate. We tested the number of parameters of the classifier and data calibrator. For digits experiment, the number of parameter of LeNet is 3.1 millions while the data calibrator has 0.18 millions of parameters, only 5.8\% compared to the model. For GTA5 to CityScapes experiments, the DRN-26 model has 20.6 millions of parameters while our data calibrator only has 0.05 millions of parameters, only 0.24\% compared to the DRN-26 model.

We thereby conclude that the proposed data calibrator is light-weight compared to the deployed model and does not bring too much overhead during deployment.

\section{Conclusion}
In summary, the proposed method not only achieves state-of -the-art performance in unsupervised domain adaptation for digits classification task and driving scene semantic segmentation task, but also be suitable for deployed models to adapt to new domains without the need to update their weights. This approach provides a feasible solution for online unsupervised domain adaptation. While the community is trying to build a monolithic model that can work across as many domains as possible, the separable approach we propose is also worth investigating.


{\small
\bibliographystyle{ieee_fullname}
\bibliography{egbib}
}




\begin{figure*}[!htp]
\centering
\includegraphics[scale=0.25]{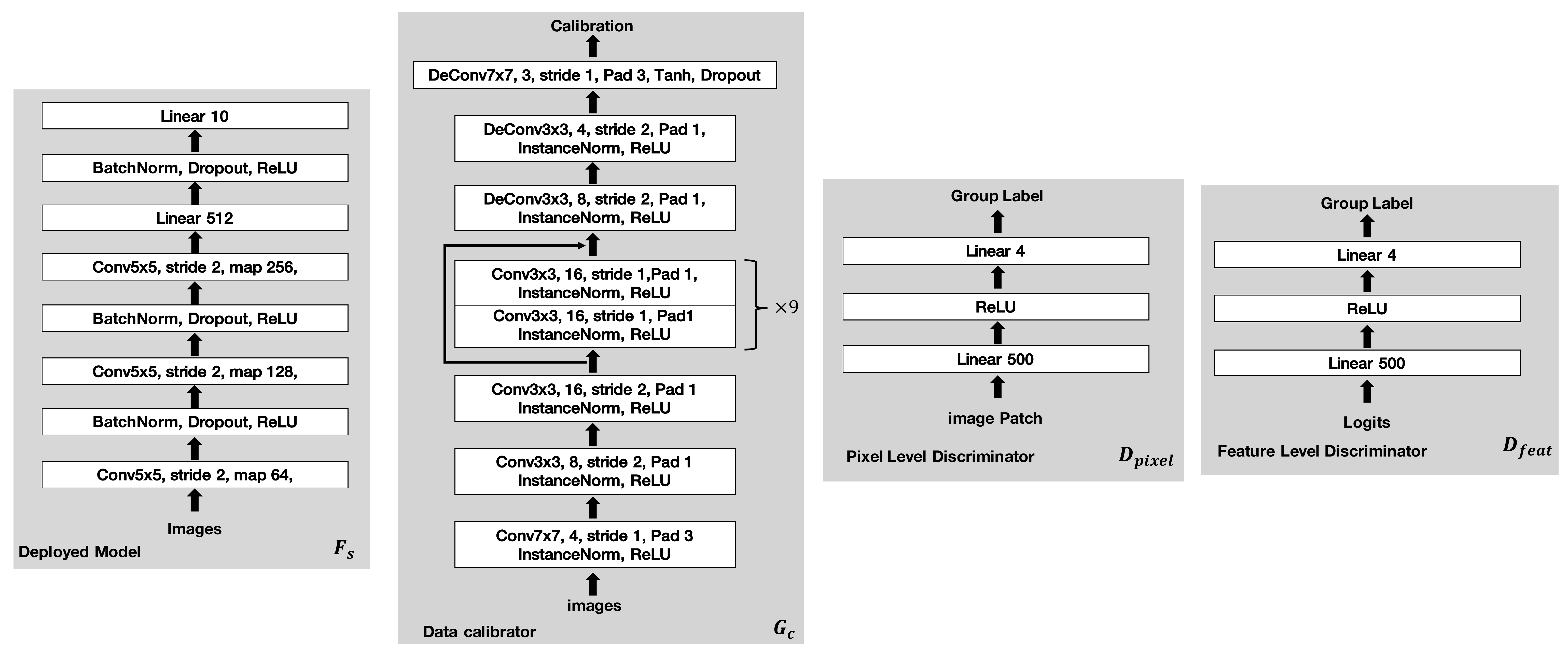}
\caption{\textbf{Network architectures used for digits experiments}~. We show the source classifier $F_s$, proposed calibrator $G_c$, pixel level domain discriminator $D_{pixel}$ and feature level domain discriminator $D_{feat}$.
}
\label{fig:architecture}
\end{figure*}

\begin{figure*}[!htp]
\centering
\includegraphics[scale=0.25]{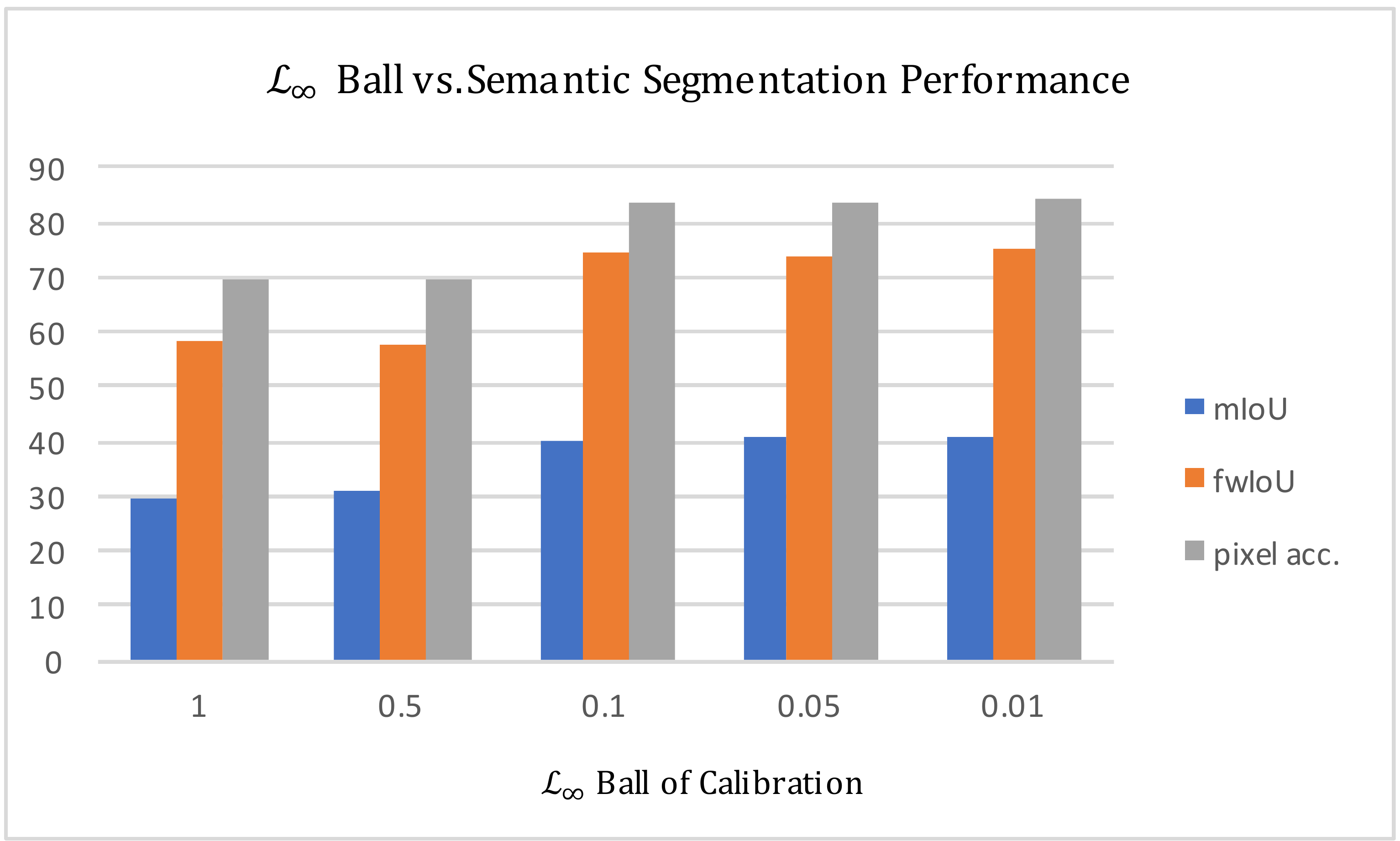}
\caption{\textbf{Performance vs. $L_{\infty}$ ball of calibration produced by the calibrator}.~ We show that with calibration that is imperceivable to human, we can achieve state-of-the-art domain adaptation performance. Calibration with large  $L_{\infty}$ ball has worse performance, probably due to overfitting or models' poor rosbutness to pixel modification in general
}
\label{fig:epsilon}
\end{figure*}


\begin{table*}[!htp]
\centering
\begin{tabular}{lll}
\hline
GTA5 to CityScapes & N. of Param.(M) & Flops(G) \\ \hline
DRN-26 & 20.6 & 200 \\
Data Calibrator & 0.05 & 2.67 \\ \hline
Digits & N. of Param.(M) & Flops(G) \\ \hline
LeNet & 3.13 & 0.03 \\
Data Calibrator & 0.18 & 0.02 \\ \hline
\end{tabular}
\caption{\textbf{Overhead of data calibrator}.~We show that our calibrator is light-weight both in terms of number of parameters and flops. Even for network as tiny as LeNet, the calibrator is small compared to it}
\label{table:overhead}
\end{table*}




\end{document}